\documentclass[letterpaper]{article} 
\usepackage{subfiles}
\usepackage{aaai2026}  
\usepackage{xcolor}
\usepackage{amsmath}
\usepackage{times}  
\usepackage{helvet}  
\usepackage{courier}  
\usepackage[hyphens]{url}  
\usepackage{graphicx} 
\usepackage{natbib}  
\usepackage{caption} 
\usepackage{algorithm}
\usepackage{algpseudocode}
\usepackage{bigstrut}
\usepackage{multirow}

\algtext*{EndIf}
\algtext*{EndFor}
\algtext*{EndWhile}

\usepackage{newfloat}
\usepackage{listings}
\DeclareCaptionStyle{ruled}{labelfont=normalfont,labelsep=colon,strut=off} 
\floatstyle{ruled}
\newfloat{listing}{tb}{lst}{}
\floatname{listing}{Listing}
\if false
\title{LiteGE: Lightweight Geodesic Embedding for Efficient Geodesics Computation and Non-Isometric Shape Correspondence}

\author{
    Written by AAAI Press Staff\textsuperscript{\rm 1}\thanks{With help from the AAAI Publications Committee.}\\
    AAAI Style Contributions by Pater Patel Schneider,
    Sunil Issar,\\
    J. Scott Penberthy,
    George Ferguson,
    Hans Guesgen,
    Francisco Cruz\equalcontrib,
    Marc Pujol-Gonzalez\equalcontrib
}
\affiliations{
    \textsuperscript{\rm 1}Association for the Advancement of Artificial Intelligence\\

    1101 Pennsylvania Ave, NW Suite 300\\
    Washington, DC 20004 USA\\
    proceedings-questions@aaai.org
}
\fi

\title{LiteGE: Lightweight Geodesic Embedding for Efficient Geodesics Computation and Non-Isometric Shape Correspondence}
\author {
    Yohanes Yudhi Adikusuma \textsuperscript{\rm 1},
    Qixing Huang \textsuperscript{\rm 1},
    Ying He \textsuperscript{\rm 2}
}
\affiliations {
    \textsuperscript{\rm 1}University of Texas at Austin
\\
    \textsuperscript{\rm 2}Nanyang Technological University \\
    quilin111@utexas.edu, huangqx@cs.utexas.edu, 
    yhe@ntu.edu.sg
}

\usepackage{bibentry}
\begin{document}

\maketitle

\begin{abstract}
Computing geodesic distances on 3D surfaces is fundamental to many tasks in 3D vision and geometry processing, with deep connections to tasks such as shape correspondence. Recent learning-based methods achieve strong performance but rely on large 3D backbones, leading to high memory usage and latency, which limit their use in interactive or resource-constrained settings. We introduce \emph{LiteGE}, a lightweight approach that constructs compact, category-aware shape descriptors by applying Principal Component Analysis (PCA) to  Unsigned Distance Field (UDFs) samples at informative voxels. This descriptor is efficient to compute and removes the need for high-capacity networks. LiteGE remains robust on sparse point clouds, supporting inputs with as few as 300 points, where prior methods fail. Extensive experiments show that LiteGE reduces memory usage and inference time by up to 300$\times$ compared to existing neural approaches. In addition, by exploiting the intrinsic relationship between geodesic distance and shape correspondence, LiteGE enables fast and accurate shape matching. Our method achieves up to 1000$\times$ speedup over state-of-the-art mesh-based approaches while maintaining comparable accuracy on non-isometric shape pairs, including evaluations on point-cloud inputs. 

\end{abstract}
\begin{links}
\link{Code}{ https://github.com/yya-111/LiteGE}
 \end{links}

\section{Introduction}
\label{sec:introduction}

Geodesic distances are a fundamental property of 3D shapes, with broad applications in deep 3D learning and digital geometry processing, such as shape matching~\cite{eisenberger2021NeuroMorph}, surface reconstruction~\cite{SPRYNSKI2008480}, parametrization~\cite{DBLP:journals/ieeemm/LeeTD05}, texture mapping~\cite{DBLP:conf/si3d/SunZZYXXH13}, segmentation~\cite{peyre2010geodesic}, and building 3D convolutional networks~\cite{masci2015geodesic}. The problem of computing geodesic distances and paths on meshes has been extensively studied for decades~\cite{mitchell1987discrete}, leading to a variety of methods, including exact algorithms~\cite{mitchell1987discrete,Chen90,Surazhsky05,Xin09,Ying13parallel,fwp,vtp} and approximate ones~\cite{FMM,Weber:2008:PAA:1409625.1409626,xin2012constant, crane2013geodesics,SVG,fDGG,DBLP:journals/tvcg/MengXTCHW22,DGG-VTP}. Despite their effectiveness, these classical methods typically incur substantial computational costs, often taking minutes or even hours to compute geodesics on large-scale models.

Recently, deep learning-based approaches, such as NeuroGF~\cite{zhang2023neurogf} and GeGNN~\cite{pang2023learning}, have demonstrated substantial efficiency improvements in geodesic computation. These methods use neural networks trained on collections of shapes to learn geodesic distances. Given a shape and query points, the network processes them to generate high-dimensional embeddings from which a lightweight multilayer perceptron (MLP) predicts the geodesic distance within milliseconds. However, the effectiveness of these methods still depends heavily on large backbone network architectures to process shapes that consume a huge amount of GPU memory. This limits their practicality in memory-constrained or interactive applications, particularly those that require geodesics in multiple shapes to be inferred at once. As such, we propose a lightweight and memory-efficient solution that supports interactive computation. Our approach is inspired by the embedding-based formulations used in NeuroGF and GeGNN. Rather than relying on large 3D neural networks, we canonicalize shapes efficiently, construct compact embeddings from them using PCA \cite{abdi2010principal} on the unsigned distance field (UDF) of the shapes \cite{chibane2020neural}, and learn geodesic distances using the embeddings. Our network, called LiteGE, preserves the generalization capability of existing methods for shapes from several categories while significantly reducing computational overhead for more than 300$\times$ in memory consumption and runtime. 



We further demonstrate the utility of LiteGE in addressing the challenging task of non-isometric shape matching, where significant geometric and topological differences complicate correspondence estimation. Although previous work, such as NeuroMorph~\cite{eisenberger2021NeuroMorph}, incorporates geodesic distances primarily as a regularization term that is not used on non-isometric shapes, our approach uses geodesic distances directly as the primary supervisory signal. Specifically, given a query point on one shape (e.g., a dog), our network identifies the corresponding point on another shape (e.g., a cat) by predicting and minimizing geodesic distances in a coarse-to-fine manner.

Shape matching in non-isometric settings is particularly challenging due to significant geometric and topological differences between shapes, especially when we need to deal with point clouds. Our approach trains a network to predict the geodesic distance between one point on the shape $X$ and one point on the shape $Y$. During training, the network is supervised with intra-shape geodesic distance, which is the geodesic distance between the two points when both lie on the same shape. At inference, a query point on 
$X$ is matched to the point on 
$Y$ that minimizes the predicted geodesic distance. This matching procedure is performed in a coarse-to-fine manner for efficiency. 

Due to the efficiency of our compact UDF representation, LiteGE supports rapid inference of correspondences for individual query points, making it particularly suitable for local region matching or sparse landmark identification on high-resolution meshes. Experiments conducted on the Skinned Multi-Animal
Linear Model (SMAL) dataset of 4-legged animals demonstrate that LiteGE achieves an accuracy comparable to the state-of-the-art Spectral-Meets-Spatial (SMS)~\cite{cao2024spectral}, which is a mesh-based method, while providing inference speeds that are 100 to 1000$\times$ faster. Notably, thanks to our novel shapes canonicalization pipeline, we can generalize the efficiency and accuracy of our method from meshes to point clouds.

Our main contributions are summarized as follows.
\begin{itemize}
\item We propose LiteGE, a lightweight geodesic prediction framework that uses PCA on unsigned distance field representations to produce compact shape embeddings.
\item Our method eliminates the need for memory-intensive 3D networks, enabling interactive and scalable geodesic computation across point clouds and meshes.
\item We show that LiteGE supports geodesic prediction on very sparse point clouds, outperforming existing methods in low-sample regimes.
\item We demonstrate that LiteGE can be applied to non-isometric shape matching for meshes and point clouds with an accuracy similar to that of state-of-the-art mesh-based methods while being up to 1000$\times$ times faster. 
\end{itemize}

\section{Related Works}
\label{sec:related-works}

\paragraph{Geodesic Computation} Traditional methods for computing geodesic distances on 3D surfaces have a long history, spanning computational geometry~\cite{mitchell1987discrete,Chen90,DBLP:journals/tvcg/XinHF12,DBLP:journals/tog/SharpC20}, partial differential equations~\cite{Sethian2000,belyaev2015,crane2013geodesics}, graph-based approaches~\cite{SVG,fDGG}, and optimization-based formulations~\cite{DBLP:journals/cad/LiuCXHLZ17,DBLP:journals/tvcg/YuanWMCXXHW23}. In contrast, deep learning-based geodesic computation is a relatively new research direction. Recent methods, such as NeuroGF~\cite{zhang2023neurogf} and GeGNN~\cite{pang2023learning}, employ neural networks to predict geodesic distances by embedding shapes and query points in high-dimensional spaces. NeuroGF constructs embeddings by concatenating global shape descriptors—computed using large 3D backbone networks like PointTransformer~\cite{zhao2021point} or MeshCNN~\cite{hanocka2019meshcnn}—with point coordinates, while GeGNN explicitly designs 3D graph neural networks specialized for geodesic tasks. However, these neural approaches typically require large memory footprints (often several gigabytes of GPU memory) even for a single geodesic query in one shape, significantly limiting their practical deployment. In this work, we observe that the shape descriptors used by NeuroGF can be constructed far more efficiently without heavy neural backbones, especially when we have a constrained set of shape categories (e.g., humans, animals, and faces). Specifically, we propose using PCA on UDF representations as a lightweight alternative. By canonicalizing shapes to a standard orientation, position, and scale, and selecting high-variance voxels near surfaces, we efficiently compute UDF-based PCA embeddings while retaining their expressiveness.

\paragraph{Shape Matching} Shape matching aims to establish correspondences between points in pairs of shapes and is frequently tackled through dense correspondence frameworks such as functional maps~\cite{DBLP:journals/tog/OvsjanikovBSBG12,litany2017deep}. Recent approaches increasingly integrate interpolation and correspondence tasks into unified neural architectures~\cite{eisenberger2021NeuroMorph,cao2023unsupervised,cao2024spectral}. However, many current methods struggle with non-isometric shape pairs, leading to increased matching errors due to substantial geometric and topological variations. Surprisingly, despite the intrinsic relationship between geodesics and shape correspondence, existing deep learning methods rarely leverage geodesic distances as a primary supervision signal. The only exception is NeuroMorph~\cite{eisenberger2021NeuroMorph}, which includes geodesic distances as a regularization term for isometric shapes but disables it when training on non-isometric shapes.
In contrast, we demonstrate that by employing geodesic distances as the main supervisory objective for shape-matching tasks, we can reduce sensitivity to non-isometric variations, enabling LiteGE to deliver superior efficiency and accuracy compared to existing state-of-the-art methods.


\begin{figure*}[h] 
    \centering     \includegraphics[width=0.85\textwidth]{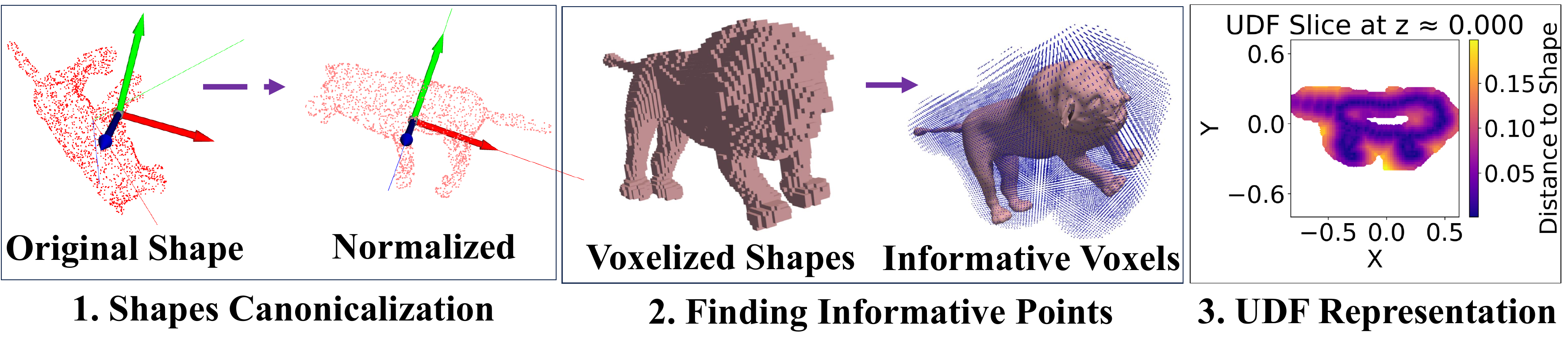} 
\caption{Illustration of shape descriptor creation process in LiteGE. First, each input point cloud is aligned to a canonical upright orientation. Next, the shape is voxelized and the informative voxels near the shapes surface are selected. Finally, we compute the unsigned distance (UDF) from the informative voxels to the shape to construct the final shape descriptor. This is as shown on the rightmost figure where we depicted the cross-section slice of the UDF heatmap through the center of the shape. }
    \label{fig:shape_rep} 
    
\end{figure*}

\section{Method}
\label{sec:method}

\subsection{Overview}
\label{subsec:overview}
This section outlines our method for constructing a lightweight yet effective geodesic representation based on PCA applied to UDF data. The objective is to generate compact shape embeddings that enable efficient geodesic distance inference and shape matching across shape categories. The overall pipeline is illustrated in Figure~\ref{fig:shape_rep}.

Our method consists of three main stages. First, we canonically transform each shape to ensure a consistent scale, orientation, and position across the dataset. Next, we construct compact shape embeddings by applying PCA to UDF data from informative points surrounding the shape. This typically results in 200- to 400-dimensional descriptors. Once the PCA-based embeddings have been computed, they are used in conjunction with a lightweight neural network to predict geodesic distances and perform fast shape matching.

\subsection{Motivations}
\label{subsec:motivation}

The key motivation behind LiteGE is that shapes within a constrained set of multiple categories share similar geometric structures. By applying shape canonicalization, which aligns scale, orientation, and position, we can significantly reduce variance across the dataset, thereby allowing the use of the PCA method to create compact yet effective shape representations. This simplification makes the learning task easier: training an MLP to separate high-dimensional, unnormalized data would require much more data and network capacity to learn complex decision boundaries. In contrast, working with normalized inputs allows the model to learn from more compact, lower-dimensional representations.

Unlike spectral methods that rely on invariant shape descriptors such as the Laplace–Beltrami Operator (LBO), which require costly eigenvalue decompositions, our approach leverages efficient normalization. This enables simpler and more flexible network architectures, leading to better scalability, faster runtime, and improved generalization on sparse or noisy inputs.




\subsection{Shape Canonicalization}
\label{subsec:canonicalization}

Our method supports geodesic inference on both meshes and point clouds. The only difference lies in the canonicalization step, which standardizes the input shape's position, orientation, and scale.
For geodesic regression, we first convert all inputs to point clouds. For shape matching, we also convert the input to point cloud, but the mesh is kept when we receive mesh as the test data. Shape matching involves two shapes, thereby increasing data variability. More precise normalization that reduces data variability can be done more efficiently using mesh data. Canonicalization is then performed in three steps: centering, scaling, and orientation. 

For centering, we shift each shape so that the point cloud sampled from it has its mean coordinate at the origin. For scaling, we adopt two strategies depending on the task: (i) Geodesic regression: We scale the point cloud such that its bounding box has a fixed surface area (e.g., 1.7 in our implementation). This method is simple and computationally efficient. (ii) Shape matching: When meshes are available, we scale each shape to have a unit surface area. For point cloud inputs, we replicate our strategy to scale meshes by making the average nearest-neighbor distance among $K$ Poisson disk samples \cite{corsini2012efficient} match that of a unit-area mesh.

Similarly, we adopt two orientation strategies: (i) Geodesic regression: We adopt PCA to find the principal components of the point cloud and align them to the global $x$, $y$, and $z$ axes. To resolve the sign ambiguity of each principal axis, we flip the axis direction (if necessary) so that it points toward the side of the shape with more point samples. The third axis is determined via the cross product of the first two to ensure a right-handed coordinate system. (ii) Shape matching: We adopt T-Net alignment since orientation consistency is important for reducing data variability when dealing with two shapes. We train a T-Net~\cite{qi2017pointnet} to predict two orthogonal 3D vectors, which are converted into a rotation matrix using Gram–Schmidt orthogonalization process~\cite{zhou2019continuity}. The network is supervised using two terms: a geodesic loss that measures the angular  difference between the predicted and  ground truth rotation matrices~\cite{alvarez2023loss}, and  an $L2$ loss that aligns the transformed point coordinates. The loss is given by:

{\fontsize{6.5}{6.5}\selectfont
\[
L_{\text{align}} = \cos^{-1} \left( \frac{\mathrm{Tr}\left(\hat{R}^\top R\right) - 1}{2} \right) +  \frac{2}{N} \sum_{i=1}^{N} \left\| \hat{p}_i - p_i \right\|_2,
\]} 

where $N$ is the number of input points, $\hat{R}$ and $R$ are the predicted and ground-truth rotation matrices, and $\hat{p}_i$ and $p_i$ are the aligned and ground-truth points coordinates.







\subsection{PCA Representation Using Informative Voxels}
\label{subsec:pca-rep}

After canonicalizing each input shape, we convert it into a voxel-based representation. Specifically, we used a binary voxel grid with a resolution of $128^3$. Each voxel is assigned a value of 1 if it lies within the shape and 0 otherwise (Figure~\ref{fig:shape_rep}, middle). In our application domain -- non-isometric shape matching involving animals and humans -- most voxels show zero or near-zero variance across the dataset, suggesting they carry little discriminative information.

To improve efficiency and focus on meaningful structures, we retain only a subset of \textit{informative voxels} whose occupancy variance exceeds a threshold $\varepsilon$. These voxels typically lie near the surface of the shape and capture the geometric variation needed to distinguish shapes effectively. 

As further note, voxelization can be done regardless whether the input is a point cloud or a mesh. But, in our experiments, the informative voxels are precomputed from available meshes dataset and frozen thereafter.

For each shape, we then compute the unsigned distance (UDF) from the center of each informative voxel to the surface, implemented efficiently via nearest-neighbor queries. The resulting UDF values form a high-dimensional vector that encodes the shape's geometry. Even with a conservative variance threshold (e.g., $\varepsilon \leq 0.0625$), each shape typically retains around 20K to 30K informative voxels, resulting in a high-dimensional representation that limits practical usage.

\begin{figure}[t] 
    \centering 
    \includegraphics[width=0.92\columnwidth]{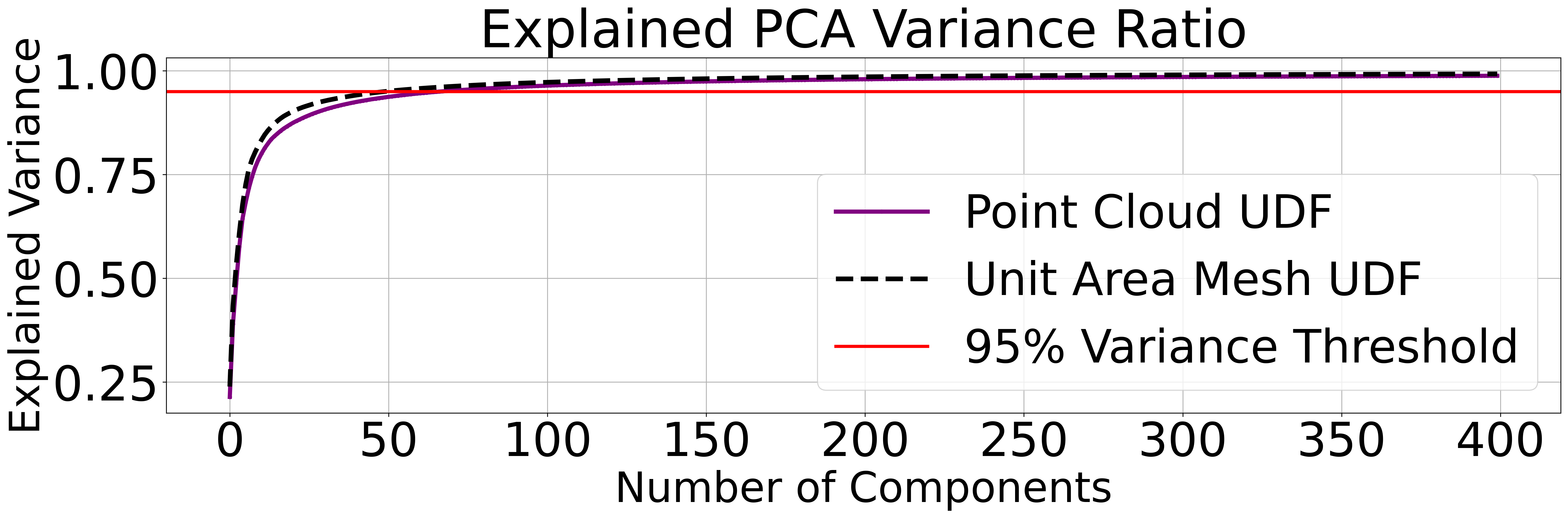} 
    \caption{Cumulative variance captured by principal components of UDFs on the SMAL dataset.``Point Cloud UDF'' denotes shape canonicalization using bounding-box scaling and PCA-based axis alignment, while ``Unit Area Mesh UDF'' refers to T-Net alignment with unit-area scaling. In both settings, the first 50 principal components account for over 95\% of the total UDF variance, highlighting the effectiveness of PCA in creating compact shape descriptors.}
    \label{fig:pca_var} 
    
\end{figure}

To reduce dimensionality while preserving geometric expressiveness, we apply principal component analysis to UDF vectors and retain only the top 50–400 components. As shown in Figure~\ref{fig:pca_var}, the first 50 principal components already capture over 95\% of the variance, confirming the strong redundancy in the full representation and justifying the effectiveness of the resulting UDF-PCA shape descriptor.

As shown in the figure, the mesh-based normalization used in shape matching achieves slightly higher explained variance than the point cloud-based normalization for geodesic regression. This is expected since mesh-based normalization employs unit-area scaling and T-Net alignment, which are more consistent than PCA axis alignment and bounding-box scaling for point clouds. PCA alignment is sensitive to axis flipping, and bounding-box scaling is less precise. This supports the observation that better normalization reduces data variability and effective dimensionality. 

\begin{figure}[t] 
    \centering 
    \includegraphics[width=0.95\columnwidth]{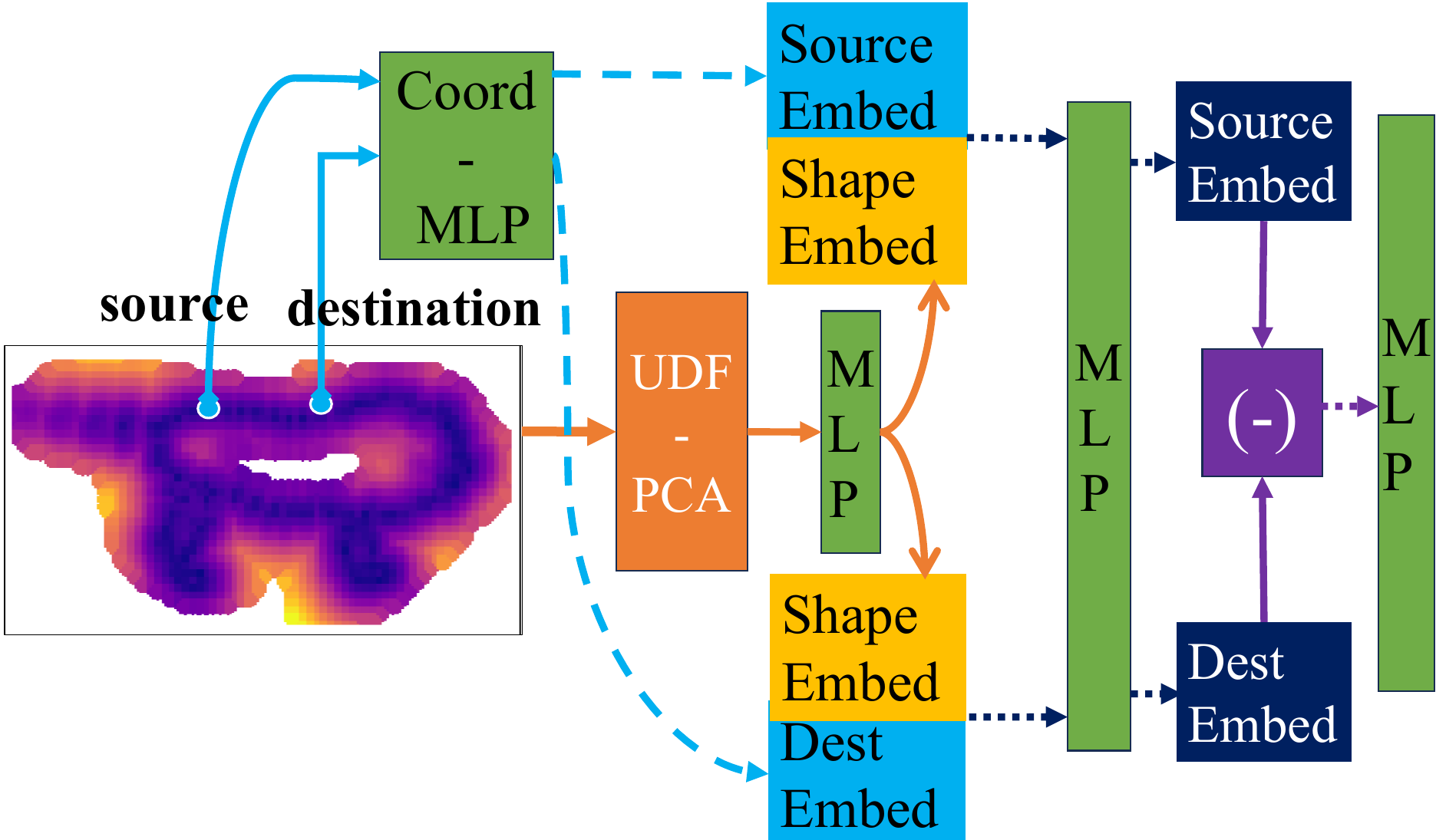} 
    \caption{Architecture of LiteGE for geodesic regression. The UDF-PCA shape descriptor and the coordinates of source and destination points are embedded using separate MLPs. These embeddings are combined to produce two shape-specific point embeddings. The geodesic distance is predicted from the difference between the point embeddings.} 
    \label{fig:net_architecture}    
\end{figure}

\subsection{Network Architecture}
\label{subsec:geodesic-net}

\paragraph{Geodesic regression}
The LiteGE architecture employs a series of lightweight multilayer perceptrons designed to predict geodesic distances between point pairs. The key idea is to embed each query point into a high-dimensional space such that the difference between two-point embeddings reflects their geodesic distance. Compared to previous methods such as GeGNN~\cite{pang2023learning} and NeuroGF~\cite{zhang2023neurogf}, our setup is significantly simpler and more memory-efficient due to the use of compact UDF-PCA shape descriptors instead of large 3D backbone networks.

As illustrated in Figure~\ref{fig:net_architecture}, the UDF-PCA shape descriptor is first passed through an MLP to produce a global shape embedding. The 3D coordinates of the source and destination points are processed through a shared MLP (Coord-MLP) to generate point-wise coordinate embeddings. Each of these is concatenated with the shape embedding and passed through another shared MLP to obtain shape-specific embeddings for the source and destination points. Their difference is then inputted into a final MLP that predicts the geodesic distance. 


\paragraph{Extension to shape matching} 
To adapt the network for shape matching, we estimate the geodesic distance between two points that lie on different shapes (e.g., shapes $X$ and $Y$), treating them as if they reside on the same surface. We modify the architecture by feeding both UDF-PCA descriptors—one from shape $X$ and one from shape $Y$—into the shared shape MLP. The corresponding query points for each shape are embedded via the Coord-MLP. Each point embedding is then concatenated with its respective shape embedding and passed through the shared-point MLP to produce shape-specified embeddings, as in the single-shape case. The final MLP predicts the average geodesic distance between the two points, measured separately in both the shape $X$ and the shape $Y$.

\paragraph{Implementation} 
The Coord-MLP, which encodes 3D point coordinates, consists of two layers, each with 200 neurons. The MLP for generating the global shape embedding from the UDF-PCA descriptor also has three layers, each containing 200 neurons. The shared MLP for producing shape-specific point embeddings contains three layers, with 200 neurons per layer. Finally, the prediction MLP, which estimates the geodesic distance from the difference between source and destination embeddings, consists of two layers with 400 neurons each. In total, LiteGE has approximately 600K parameters, only 10\% of GeGNN, signifying its compact and efficient design.

\paragraph{Training}
 Ground-truth geodesic distances are generated using FastDGG~\cite{fDGG}, with accuracy thresholds of 0.3\% for training and 0.1\% for testing. Here, $L1$ loss is used to train LiteGE networks. 

\subsection{Inference}
\label{subsec:geo_infer}

During inference, each shape is first canonicalized, as described previously. Then, our network requires two inputs: (1) the UDF-PCA representation of a canonicalized shape, and (2) the 3D coordinates of the query points (e.g., source and destination). 

To construct the UDF-PCA descriptor, we first compute the unsigned distances by measuring the distance from each informative voxel center to the shape surface using nearest-neighbor queries. This UDF vector is then projected onto the principal component basis, resulting in a compact representation. Specifically, the $i$-th component of the UDF-PCA vector is computed as:

{\fontsize{8}{8}\selectfont
\[
z_{\text{UDF-PCA},i} = \left\langle \mathbf{x}_{\text{UDF}} - \mathbf{\bar{x}}_{\text{UDF}},\ \mathbf{p}_{\text{UDF},i} \right\rangle,
\]}  

where $\mathbf{x}_{\text{UDF}}$ is the UDF vector of the current shape, $\mathbf{\bar{x}}_{\text{UDF}}$ is the mean UDF vector computed from the training set, and $\mathbf{p}_{\text{UDF},i}$ is the $i$-th principal component.

When handling multiple queries on a single shape, we can precompute the shape embedding for reuse. If certain query points appear multiple times, their shape-specific embeddings can also be cached. The same optimization applies in shape matching, where embeddings of both input shapes can be precomputed once and reused across multiple queries.

\subsection{Shape Correspondence}
\label{subsec:shapematch_infer}
We perform shape matching using a coarse-to-fine strategy, as detailed in Algorithms~\ref{algo:cache_building} and \ref{algo:point_matching}. Given a query point $Q_{X}$ on shape $X$, the algorithm efficiently identifies its best match on shape $Y$ by iteratively refining the predictions, starting from a sparse set of candidate points. 

The process begins by constructing a multi-tier nearest-neighbor cache on shape $Y$ (Algorithm~\ref{algo:cache_building}). We define $K$ tiers of point sets, where Tier 1 is the sparsest (e.g., $N_1$ points), and Tier $K$ includes all mesh vertices. For each point in Tier~$i$, we store its $M_i$ nearest neighbors in Tier~$i{+}1$.

Using the cache, Algorithm~\ref{algo:point_matching} finds a match for $Q_X$ by sequentially querying the geodesic prediction network $G$ at increasing resolution. In each tier, we select the best match from the current candidate set and use the cache to retrieve a set of neighbors for the next tier. This iterative process continues until Tier~$K$, where the final match is returned.

This coarse-to-fine strategy significantly reduces the number of geodesic queries required, enabling efficient matching even on high-resolution meshes. In our implementation, we use three cache tiers, each created by randomly sampling mesh vertices. Both NN-cache construction and query operations are parallelized using \texttt{pynanoflann} and Pytorch.

\begin{algorithm}[htb!]
\caption{Multi-Tiered Nearest Neighbor Caching}
\label{algo:cache_building}
\begin{algorithmic}[1]
\Require Shape $\mathcal{S}$, multi-tiered point sets $\{P_i(\mathcal{S})\}_{i=1}^K$, where $P_i(\mathcal{S})$ contains $N_i$ sampled points and $N_1 < \cdots < N_K$ (with $N_K$ being all vertices); neighbor counts $M_1, \ldots, M_{K-1}$
\Ensure $K$-tiered Nearest Neighbor Cache $\mathcal{C}$ for shape $\mathcal{S}$
    \State Initialize $\mathcal{C}(S) \gets \emptyset$
    \For{$i = 1$ \textbf{to} $K-1$} \Comment{Construct cache between adjacent tiers}
        \For{each point $p \in P_i(\mathcal{S})$}
            \State Find $M_i$ nearest neighbors  of $p$ in $P_{i+1}(\mathcal{S})$ using Euclidean distance.
            \State Store the mapping $p\rightarrow  \mathcal{N}_{p,i+1}$ in $\mathcal{C}(S)$
        \EndFor
    \EndFor

\State \Return $\mathcal{C(\mathcal{S})}$

\end{algorithmic}
\end{algorithm}

\begin{algorithm}[htb!]
\caption{Coarse-to-Fine Shape Matching Strategy}
\label{algo:point_matching}
\begin{algorithmic}[1]
\Require Query point $q_X$ on shape $X$, nearest neighbor cache $\mathcal{C}(Y)$ for shape $Y$, geodesic inference network $G$, UDF-PCA descriptors  $\mathbf{z}_{\text{pca}}(X)$ and $\mathbf{z}_{\text{pca}}(Y)$, multi-tiered point sets $\{P_i(Y)\}_{i=1}^K$
\Ensure Corresponding point $q_Y$ on shape $Y$

\State Initialize candidate set $\mathcal{C}_{\mathrm{cand}} \gets P_1(Y)$ 

\For{$i = 1$ \textbf{to} $K$} \Comment{Iterate through tiers}
    \State$q_{i}(Y) \gets \arg\min_{p \in \mathcal{C}_{\text{cand}}} G(q_X, p, \mathbf{z}_{\text{pca}}(X), \mathbf{z}_{\text{pca}}(Y))$ \Comment{Query for best match in the current candidates}

     \If{$i < K$} \Comment{If not the finest tier}
        \State $\mathcal{C}_{\text{cand}} \gets $ the $M_i$ nearest neighbors of $q_{i}(Y)$ from $P_{i+1}(Y)$ using $\mathcal{C}(Y)$ \Comment{Refine search space}
    \Else \Comment{Reach the finest tier}
        \State $q_Y \gets q_{K}(Y)$ \Comment{Final match found}
    \EndIf
\EndFor
\State \Return $q_Y$

\end{algorithmic}
\end{algorithm}

\subsection{Geodesic Path Tracing}
\label{subsec:path-retrieval}

To recover a geodesic path between two points, we use a gradient-based tracing method similar to GeGNN~\cite{pang2023learning}. The key idea is to iteratively update the destination point along the negative gradient of the predicted geodesic distance until convergence.

Let $\mathbf{s}$ denote the source point, $\mathbf{d}^{(t)}$ the destination point at iteration $t$, and $\mathbf{z_{\text{UDF-PCA}}}$ the shape descriptor. The geodesic prediction $G(\mathbf{s}, \mathbf{d}^{(t)}, \mathbf{z_{\text{UDF-PCA}}})$ is differentiated with respect to $\mathbf{d}^{(t)}$ to obtain the gradient, which guides the update:

{\fontsize{8}{8}\selectfont
$$
\mathbf{d}^{(t+1)} = \text{Proj}\left( \mathbf{d}^{(t)} - \eta \nabla_{\mathbf{d}} G(\mathbf{s}, \mathbf{d}^{(t)}, \mathbf{z}_{\text{UDF-PCA}}) \right),
$$}

where $\eta$ is the learning rate and $\text{Proj}(\cdot)$ denotes projection onto the shape (e.g., via nearest-neighbor search in the point cloud). The process terminates when the predicted distance ceases to decrease or falls below a threshold $\varepsilon$.

If a mesh is available, the resulting path, represented by a sequence of intermediate points, can be projected onto dense mesh samples to ensure that it lies on the surface.

\section{Results and Discussions}
\label{sec:results}

\begin{table}[htb!]
  \centering
  \setlength{\tabcolsep}{1mm}
  \small
    \begin{tabular}{r|r|r|r|r}
      \hline
      \multirow{2}{*}{$N_Q$} & \multicolumn{2}{c|}{Memory (MB)} & \multicolumn{2}{c}{Inference Time (ms)} \\
      \cline{2-5}
      & NeuroGF & LiteGE & NeuroGF & LiteGE \\
      \hline
      \hline
      1  & 162    & 1 \textcolor{orange}{(162$\times$)}        & 88   & 1.9 \textcolor{orange}{(47$\times$)} \\
      4  & 647    & 2.7 \textcolor{orange}{(240$\times$)}      & 107  & 2 \textcolor{orange}{(54$\times$)} \\
      8  & 1294   & 4.5 \textcolor{orange}{(288$\times$)}      & 145  & 2.6 \textcolor{orange}{(56$\times$)} \\
      32 & 5172   & 17.25 \textcolor{orange}{(296$\times$)}    & 763  & 7.3 \textcolor{orange}{(104$\times$)} \\
      64 & 10344  & 34.5 \textcolor{orange}{(300$\times$)}     & 1529 & 13.4 \textcolor{orange}{(114$\times$)} \\
      96 & OOM    & 51.8                                & OOM  & 19.1 \\
      \hline
    \end{tabular}
    \caption{Comparison of LiteGE with NeuroGF in terms of memory consumption and runtime performance for varying numbers of queries. Reported runtimes include all necessary preprocessing for both methods. Each point cloud receives a single query, and $N_Q$ denotes the total number of distinct queries evaluated. All tests are conducted on 2K point clouds. Improvements of LiteGE over NeuroGF are highlighted in orange. ``OOM'' indicates out-of-memory failure.}
  
  \label{tab:runtime_mem_ngf_ours}
\end{table}

 All experiments were conducted on a dedicated GPU server equipped with an NVIDIA RTX A4000 GPU, an Intel Xeon Gold 5315Y CPU, and 45 GB of RAM. We evaluate the efficiency and effectiveness of LiteGE across three key tasks: geodesic distance prediction, non-isometric shape matching, and geodesic path tracing. We begin by assessing LiteGE’s core capability in predicting geodesic distances, then demonstrate its application to downstream tasks. Due to space constraints, additional implementation details, ablation, and extended experiments are provided in the appendix.

\subsection{Lightweight Geodesic Distance Inference}
\label{subsec:lightweightinference}

We evaluate LiteGE against NeuroGF~\cite{zhang2023neurogf} using the official codes and pretrained model provided by the authors. Both methods are tested on point cloud inputs, with shape canonicalization performed as described in the Methods section. NeuroGF uses a DGCNN~\cite{wang2019dynamic} backbone, a deep graph convolutional network with a large multi-layer architecture (5–6 layers), which results in high memory usage and inference costs. In contrast, LiteGE employs a compact MLP-based architecture and uses UDF-PCA shape descriptors, significantly reducing computational costs. Since GeGNN~\cite{pang2023learning} does not support point cloud input, we exclude it from comparisons. However, we note that GeGNN and NeuroGF exhibit similar accuracy and runtime performance when evaluated on mesh inputs, as reported in their respective papers.

For benchmarking, we use the SMAL PCA model~\cite{SMALCVPR:2017} to generate 12,000 diverse quadruped animal meshes. 
All reported errors are normalized such that the ground-truth geodesic distances have a mean value of 100.0. We exclude the memory cost of storing the UDF-PCA basis, as it remains constant (10–20 MB) regardless of the number of queried point clouds.

\begin{figure}[t] 
    \centering 
    
    \includegraphics[width=0.98\columnwidth]{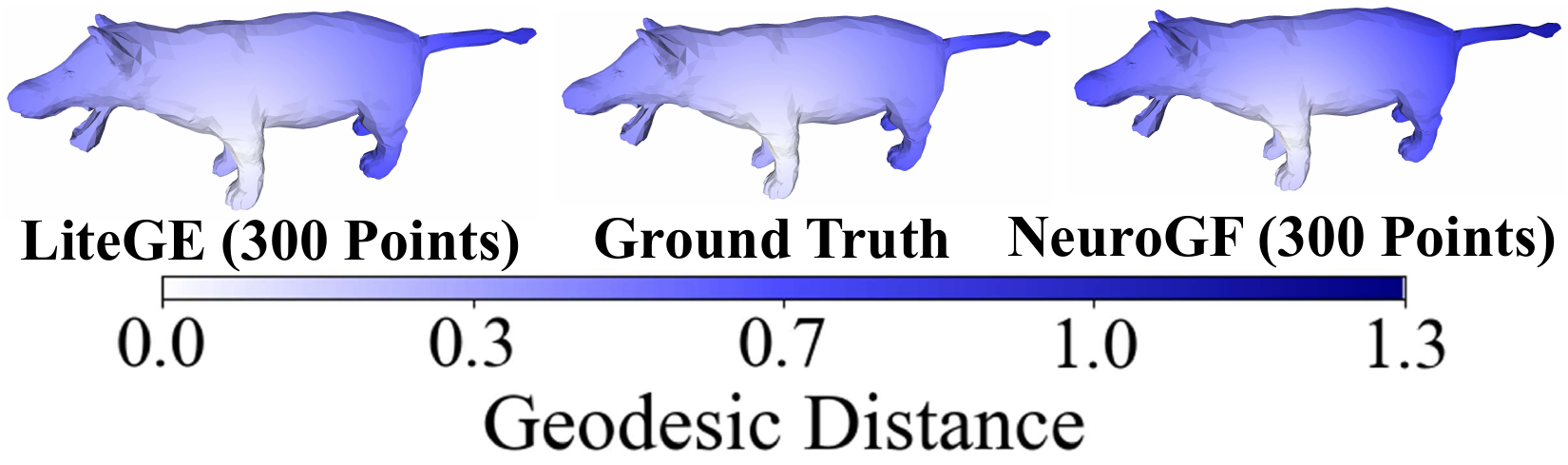} %
\caption{Visual comparison of geodesic distance predictions on a sparse (300-point) pig model, with the source point located on the left front paw. Distances are visualized using a heatmap: warmer colors indicate larger distances, while cooler colors indicate smaller ones. The mean prediction L1-errors are 24 for NeuroGF and 2.5 for LiteGE, with ground-truth distances normalized to a mean of 100. This highlights LiteGE's accuracy on sparse point clouds.}
    \label{fig:visualgeod} 
\end{figure}

 As shown in Table \ref{tab:runtime_mem_ngf_ours}, LiteGE is up to 300$\times$ more memory-efficient and over 100$\times$ faster than NeuroGF. Moreover, NeuroGF shows significantly lower accuracy across our SMAL test cases (Table \ref{tab:litege_comparison}), indicating limited capacity to generalize to unseen shape categories. In contrast, LiteGE remains robust even on sparse point clouds with only 300 samples (Figure \ref{fig:visualgeod}), where NeuroGF fails with a 75 L1 error. This is likely because unsigned distances from informative voxels in LiteGE remain stable across sampling densities.

Also, LiteGE is robust to shapes with noise and missing regions. With the ground-truth distances normalized to a mean of 100, given 2K-point clouds sampled from unit area meshes in the SMAL dataset, LiteGE has a mean L1 error of 4.3 when 0.01-scale Gaussian noise is added to the point clouds. Moreover, LiteGE achieves mean L1 errors of 3.7 and 4.5 when 10\% and 15\% of the points are randomly removed by two slicing planes.
 
 Finally, in Table \ref{tab:litege_comparison}, LiteGE demonstrates strong generalization to unseen and diverse datasets. When trained on a combined dataset of 8K meshes from the SMAL model~\cite{SMALCVPR:2017} and 8K from the SURREAL human dataset~\cite{groueix20183d}, LiteGE generalizes to the FAUST  dataset ~\cite{bogo2014faust}, even when tested using sparse inputs (700 points). Furthermore, when trained on 3K models from the Objaverse-XL \cite{deitke2023objaverse} dataset and tested on a separate validation set, LiteGE generalizes across diverse shape categories. 


 \if false
\begin{table}[htb!]
  \centering
  \caption{NeuroGF and LiteGE memory and runtime comparison given different numbers of point clouds for queries.  Each point cloud is queried with 1 geodesic distance query.  Number of distinct queries is given in first column. LiteGE improvements compared to NeuroGF is highlighted in orange. Point clouds with 2K samples are used to test.}
  \begin{adjustbox}{width=0.8\columnwidth} 
   \begin{tabularx}{\linewidth}{|r|X|X|X|p{1.7cm}|}
    \hline
    \multirow{2}{*}{\textbf{Q}} & \multicolumn{2}{c|}{\textbf{Memory (MB)}} & \multicolumn{2}{c|}{\textbf{Inference Time (ms)}} \\
    \cline{2-5}
    & $M_{ngf}$ & $M_{LiteGE}$ & $T_{ngf}$ & $T_{LiteGE}$ \\
    \hline
    1 &
      162 &
      \multicolumn{1}{l|}{1 \textcolor{orange}{(162×)}} &
      88 &
      1.9 \textcolor{orange}{(47x)}
      \bigstrut\\
    \hline
    4 &
      647 &
      \multicolumn{1}{l|}{2.7 \textcolor{orange}{(240x)}} &
      107 &
      2 \textcolor{orange}{(54x)}
      \bigstrut\\
    \hline
    8 &
      1294 &
      \multicolumn{1}{l|}{4.5 \textcolor{orange}{(288x)}} &
      145 &
      2.6 \textcolor{orange}{(56x)}
      \bigstrut\\
    \hline
    32 &
      5172 &
      \multicolumn{1}{l|}{17.25 \textcolor{orange}{(296x)}} &
      763 &
      7.3 \textcolor{orange}{(104x)}
      \bigstrut\\
    \hline
    64 &
      10344 &
      \multicolumn{1}{l|}{34.5 \textcolor{orange}{(300x)}} &
      1529 &
      13.4 \textcolor{orange}{(114x)}
      \bigstrut\\
    \hline
    96 &
      OOM &
      51.8 &
      OOM &
      19.1
      \bigstrut\\
    \hline
    \end{tabularx}%
    \end{adjustbox} 
  \label{tab:runtime_mem_ngf_ours}%
\end{table}%
\fi

\begin{table}[t]
  \centering

  
    \centering
    
    \small
    \setlength{\tabcolsep}{1mm}
    \begin{tabular}{|c|c|c|c|c|}
    \hline
     LiteGE Train Data & Test Data & \multicolumn{1}{l|}{ \#Samples} & $\varepsilon_{LGE}$ & $\varepsilon_{NGF}$ \\
    \hline\hline
  SMAL & SMAL & 300 & 3.4 & 75 \\
     SMAL & SMAL & 2000 & 2.3 & 13 \\
     SMAL \& SURREAL & FAUST &  700 & 2.6 &  NA \\
     Objaverse-XL & Objaverse-XL & 2000 & 3.9 &  NA \\
    
    \hline
    \end{tabular}
    

  \caption{
  Comparison of mean L1 errors of LiteGE ($\varepsilon_{LGE}$)  and NeuroGF ($\varepsilon_{NGF}$) on the SMAL dataset under varying point-cloud sample densities, as well as LiteGE’s generalization ability (measured as median L1 errors) on unseen (FAUST) or diverse (Objaverse) datasets. Notably, NeuroGF fails on sparse point clouds with 300 samples, with an L1-error of 75. Ground-truth distances are normalized to have a mean of 100.  }
  \label{tab:litege_comparison}
\end{table}


\begin{table}[htb!]
  \centering
  \setlength{\tabcolsep}{1mm}
  \small
    \begin{tabular}{c|c|c|c|c|c|c|c|c}
    \hline
    \multicolumn{2}{c|}{} &
      \multicolumn{3}{c|}{SMS} &
      \multicolumn{3}{c}{LiteGE } 
      \\
    \hline
    \textbf{Mesh Type} &
      $|V|$ &
      \multicolumn{1}{l|}{T (ms)} &
      \multicolumn{1}{l|}{AUC} &
      \multicolumn{1}{l|}{Error} &
      \multicolumn{1}{l|}{T (ms)} &
      \multicolumn{1}{l|}{AUC} &
      \multicolumn{1}{l}{Error} 
      \\
      \hline
    
    \hline
    Template &
      4K &
      21300 & 
      79.4 &
      2.2 &
      185 &
      79.3 &
      2.5
      %
      \\
    \hline
    Remeshed &
      5K &
      27000 &
      73.1 &
      7.44 &
      119 &
      74 &
      7.2 
      \\
    \hline

    Anisotropic &
      10K &
      38500 &
      74.9 &
      6.8 &
      120 &
      74.3 &
      7 
      \bigstrut\\
    \hline
    
    Broken Mesh & 5K & 25700 & 33.2 & 28 & 185 & 69.5 & 8 
    \\
  \cline{1-8}
    Point Cloud & 8K   & \multicolumn{3}{c|}{\multirow{2}{*}{NA}} & 200 & 71.5 & 7.6 
    \\
  \cline{1-2} \cline{6-8}
  Point Cloud & 500 & \multicolumn{3}{c|}{}                     & 126 & 69.2 & 8.1 \\
  \hline

    \end{tabular}%
    \caption{Non-isometric shape correspondence results. We query 1K points matches using LiteGE and report mean geodesic error scaled by 100 on unit-area meshes. For template models, we use a threshold of 0.1 to compute the AUC; for all other settings, a threshold of 0.2 is used. Moreover, for template models, an ensemble of 2 LiteGE models is used to test, while no ensemble is used on other tests. Here, SMS runtime includes both the LBO precomputation and final query time. The LiteGE runtime includes UDF-PCA generation, NN-cache construction, and final query time. }
    
  \label{tab:shapematch}%
\end{table}%

\subsection{Fast and Accurate Shape Matching} \label{subsec:results-shape-matching}

Table~\ref{tab:shapematch} presents the results of our shape matching experiments on the 12K SMAL meshes used for geodesic inference evaluation. For both training and testing, we used randomly rotated meshes that were then canonicalized using T-Net and our mesh normalization pipeline. To ensure reliable supervision, we only included meshes with alignment error of less than 0.66 radians during training. To stabilize training, we repeated the first training epoch eight times and selected the best model checkpoint before continuing training.  

We compare our method with Spectral-Meets-Spatial (SMS)~\cite{cao2024spectral}, a recent state-of-the-art approach. Due to disk space constraints for storing LBO parameters, we trained SMS on 6K meshes. However, as shown in the appendix, the evaluation accuracy of SMS saturates after using more than 1200 meshes. SMS was trained using its official training schedule with the non-isometric setting enabled, sampling 6400 random shape pairs per epoch.

For LiteGE, we used 1000 query points to ensure good coverage of the target mesh, particularly for high-resolution inputs. The configuration of our multi-tiered NN-cache is detailed in the appendix. We evaluated both methods on a variety of test inputs, including (1) SMAL template meshes, (2) isotropic and anisotropic remeshed models, (3) broken meshes with 40\% of faces randomly removed, and (4) point clouds with either 8K or 500 samples. For remeshed models, ground-truth correspondences were obtained via barycentric projection to template meshes. Note that we evaluate only across distinct shapes; matching identical shapes (once uniformly scaled, centered, and PCA-aligned) is trivial.



For broken meshes and point clouds, we adopt a canonicalization strategy that scales the shape so that the average nearest-neighbor distance of $K$ Poisson-disk samples matches that of a unit area mesh.



As shown in Table~\ref{tab:shapematch}, LiteGE achieves an accuracy comparable to SMS on clean meshes, while being up to 300$\times$ faster when matching 1K points. On the 5K-vertex remeshed model, LiteGE computes the UDF-PCA descriptors and NN-cache in 13 ms before performing per-point matching. Thus, it is up to 1000$\times$ faster than SMS when matching 250 points or less in this setting. On remeshed models as well, arguably a more realistic test case, LiteGE generalizes better thanks to its UDF-based representation, outperforming SMS in accuracy, as shown in Figure~\ref{fig:shapematch_illustrate}. We also show in the appendix that augmenting SMS training with remeshed shapes does not improve its generalization. Furthermore, SMS fails in broken meshes (with error $>$ 25), while LiteGE maintains strong performance. LiteGE also generalizes well to dense and sparse point clouds, achieving mesh-level accuracy even on sparse inputs with just 500 points (Figure~\ref{fig:shapematch_illustrate}).


\begin{figure}[t]
    \centering
  
\begin{minipage}{0.298\columnwidth}
         \centering
         \includegraphics[width=\linewidth]{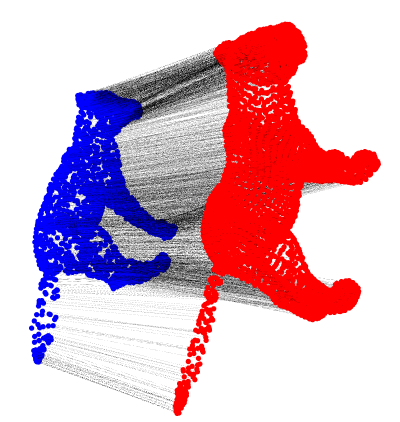}
     \end{minipage}\hspace{0.05in}
     \begin{minipage}{0.57\columnwidth}
         \centering
         \includegraphics[width=\linewidth]{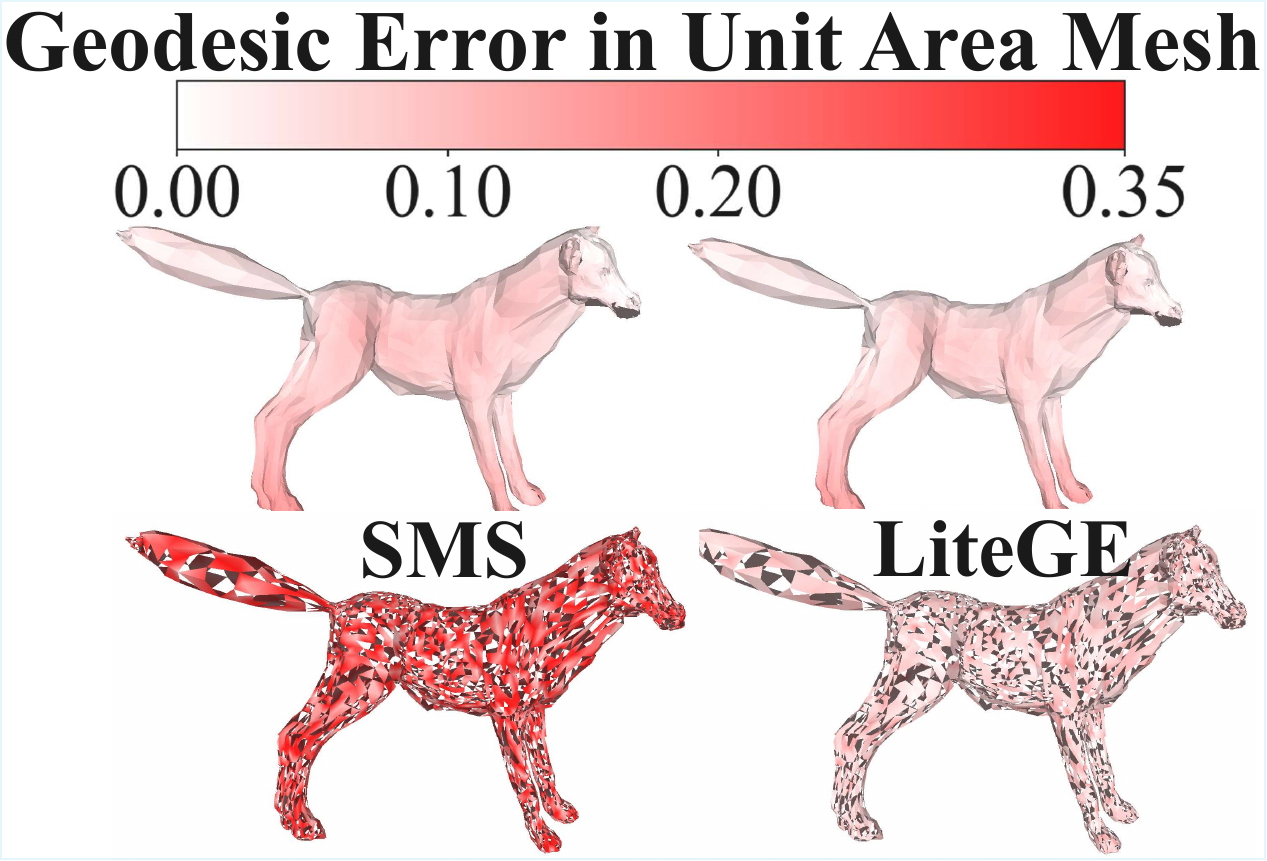}
     \end{minipage}

  \caption{Visual results on shape matching. Left: LiteGE accurately performs point cloud matching on an 8K-sample point cloud derived from a unit-area mesh, achieving a mean geodesic error of 0.022. Matched point pairs are connected by lines. Right: Shape matching performance of LiteGE and SMS on remeshed 5K-vertex models and their broken variants (with 40\% of faces randomly removed). Average errors across all test shape pairs are visualized on a dog model. }
    \label{fig:shapematch_illustrate}
\end{figure}

\begin{table}[h]
  \centering
  \setlength{\tabcolsep}{1mm}
    \small
      \begin{tabular}{r|r|r|r}
        \hline
        \#PC Samples & Error & Memory (MB) & Learning rate \\
        \hline
        \hline
        800   & 4.73 & 4.4  & 0.66 \\
        1,000  & 4.27 & 4.22 & 0.66 \\
        8,000  & 3.77 & 5.1  & 0.35 \\
        15,000 & 3.66 & 6.4  & 0.25 \\
        \hline
      \end{tabular}
      

  \caption{Geodesic path tracing results on a batch of eight shapes randomly chosen from the SMAL dataset. We report the median prediction error, with ground truth distances scaled such that their mean is 100. The learning rate used for each example is listed in the last column. }
  \label{fig:geotrace_and_path}
\end{table}

\subsection{Geodesic Paths Tracing}
\label{subsec:results-path-tracing}
In addition to geodesic distance inference, LiteGE also supports geodesic path tracing via gradient-based backtracking. As shown in Table~\ref{fig:geotrace_and_path}, LiteGE enables accurate path reconstruction on a batch of eight sparse point clouds, each containing fewer than 1,000 points. Remarkably, this is achieved with a memory footprint of under 5~MB.

\section{Conclusion and Future Directions}
\label{sec:conclusion}
We introduced LiteGE, a lightweight neural framework for geodesic distance prediction on 3D shapes, based on PCA embeddings of unsigned distance fields. Unlike prior neural methods that rely on large, memory-intensive backbones, LiteGE offers a highly efficient and scalable solution that supports both meshes and point clouds, including sparse, noisy or incomplete inputs. Extensive experiments show that LiteGE maintains an accuracy comparable to or superior to that of state-of-the-art methods while achieving over 300$\times$ reductions in memory and runtime. Moreover, we demonstrated its effectiveness in non-isometric shape matching, where LiteGE achieves up to 1000$\times$ speedup over existing methods and remains robust under sparse inputs.

Future work includes extending LiteGE to produce geodesic-aware shape descriptors for other downstream tasks, such as shape segmentation, analysis, and retrieval. Furthermore, incorporating temporal consistency for dynamic or time-varying shapes would further expand the practical utility of LiteGE in real-world 4D applications. Finally, exploring self-supervised or few-shot variants could reduce dependence on labeled data, improving generalization across diverse shape categories.
\section*{Acknowledgments} This project was partially supported by the NSF-IIS-2413161 grant and the Ministry of Education, Singapore, under its Academic Research Fund Grant (RT19/22).

\bibliography{aaai2026}
\if false
\section{Reproducibility Checklist}

This paper: \\
\begin{itemize}

\item Includes a conceptual outline and/or pseudocode description of AI methods introduced. \textbf{Yes.}

\item Clearly delineates statements that are opinions, hypothesis, and speculation from objective facts and results. \textbf{Yes.}

\item Provides well marked pedagogical references for less-familiare readers to gain background necessary to replicate the paper. \textbf{Yes.} References to PCA and UDF concepts are included for less familiar readers.

\end{itemize}

Does this paper make theoretical contributions? \textbf{No.}
\\ \\
Does this paper rely on one or more datasets? \textbf{Yes.}
\\ 
\begin{itemize}
  \item A motivation is given for why the experiments are conducted on the selected datasets. \\
  \textbf{Yes.} We provided explanation in the supplementary materials.

  \item All novel datasets introduced in this paper are included in a data appendix. \\
  \textbf{N/A}

  \item All novel datasets introduced in this paper will be made publicly available upon publication of the paper with a license that allows free usage for research purposes. \\
  \textbf{N/A}

  \item All datasets drawn from the existing literature (potentially including authors’ own previously published work) are accompanied by appropriate citations. \\
  \textbf{Yes.}

  \item All datasets drawn from the existing literature (potentially including authors’ own previously published work) are publicly available. \\
  \textbf{Partial.} Both SMAL PCA model and SURREAL dataset are publicly available. We will share the specific 12K meshes sampled from the SMAL dataset in the Appendix.

  \item All datasets that are not publicly available are described in detail, with explanation why publicly available alternatives are not scientifically satisficing. \\
  \textbf{Yes.} There are no publicly available datasets on geodesic distance ground truth computed from the SMAL, SURREAL, or FAUST datasets used here, but we provided detailed explanation on how we created them in the Appendix. We will make these geodesics ground truths data public when our paper gets accepted.
  
\end{itemize}

  Does this paper include computational experiments? \\
  
  \textbf{Yes.} If yes, please complete the list below.
\\ 
  \begin{itemize}
    \item This paper states the number and range of values tried per (hyper-)parameter during development of the paper, along with the criterion used for selecting the final parameter setting. \\
    \textbf{Partial.} We reported how the number of PCA dimensions used affect our error rate and explained variance ratio as we think it is an important detail. Some other parameters were adjusted during development for practical reasons, but were not systematically recorded, as they are not central to the novelty or conclusions of this work.

    \item Any code required for pre-processing data is included in the appendix. \\
    \textbf{Partial.} We provided the code to generate geodesic distance ground truth for part of our dataset. It can be easily modified to generate geodesic ground truth for other datasets. We did not have time to organize all of the dataset creation codes.

    \item All source code required for conducting and analyzing the experiments is included in a code appendix. \\
    \textbf{Yes.}

    \item All source code required for conducting and analyzing the experiments will be made publicly available upon publication of the paper with a license that allows free usage for research purposes. \\
    \textbf{Yes.}

    \item All source code implementing new methods have comments detailing the implementation, with references to the paper where each step comes from. \\
    \textbf{Partial,} due to time constraints. We will comment all of the source code before publication.

    \item If an algorithm depends on randomness, then the method used for setting seeds is described in a way sufficient to allow replication of results. \\
    \textbf{No.} The randomness used in our algorithm does not significantly affect the results. We can reproduce the error figures in the paper within 5\% tolerance.

    \item This paper specifies the computing infrastructure used for running experiments (hardware and software), including GPU/CPU models; amount of memory; operating system; names and versions of relevant software libraries and frameworks. \\
    \textbf{Yes.}

    \item This paper formally describes evaluation metrics used and explains the motivation for choosing these metrics. \\
    \textbf{Yes.}

    \item This paper states the number of algorithm runs used to compute each reported result. \\
    \textbf{Yes.} We state the number of random test samples in the Appendix.

    \item Analysis of experiments goes beyond single-dimensional summaries of performance (e.g., average; median) to include measures of variation, confidence, or other distributional information. \\
    \textbf{Yes.} We include AUC measures.

    \item The significance of any improvement or decrease in performance is judged using appropriate statistical tests. \\
    \textbf{No.} Our experiments showed consistent results across different runs and datasets. Given the observed stability, we did not perform formal statistical significance testing.

    \item This paper lists all final (hyper-)parameters used for each model/algorithm in the paper’s experiments. \\
    \textbf{Yes.}
  
\end{itemize}
\fi

\end{document}


\maketitle

In this supplementary material, we present ablation studies, training parameters and additional experimental results and notes on LiteGE. This document is organized as follows. 

First, LiteGE is trained to predict the difference between the geodesic and Euclidean (straight-line) distances. So, we presented the ablations experiments on this in the first section dedicated for geodesic regressions. We also present training parameters and experimental results related to LiteGE regression networks there. Other than this, we focused on how different number of PCA dimensions can influence the error. We also presented our geodesic dataset details after it in the same section. 

Then, we will present the shape matching training and NN-cache configuration details in the second section. We also presented ablation results on shape matching here, for the reason why we used the TNet \cite{qi2017pointnet} alignment method, why we filter the TNet output for training and why we launch 8 training passes each time. Finally, in the last section, we presented SMS \cite{cao2024spectral} results when training with different number of meshes or when training it using randomly remeshed models as data augmentation.

It is important to note here that the error figure in our experiments are all estimated using random samples to save time. Specifically, for testing SMS, we sample 500 meshes pairs randomly, from which we can calculate the mean geodesic error and AUC scores. For testing LiteGE shape matching result, we used approximately 1 - 5 million random vertex matches from random meshes pairs in the testing set to estimate the mean geodesic error and AUC scores. Finally, for geodesics regression networks, we used around 2 million random geodesics distances between random sources and destinations pairs on random shapes in our testing set to calculate the error.

\section{Lightweight Geodesic Inference Training Details and Ablations} 
\subsection{Training Details and Ablations}
We trained the geodesic inference network with L1 loss using Adams optimizer with 10 epochs and batch size 3072, with early stopping applied. We used cosine annealing learning rate schedule that starts from $10^{-2}$ and ends with $10^{-4}$ learning rate. We trained the network to predict the difference between straight line distance between 2 points and the geodesic distance. This shows to improve accuracy by 5-7\% as seen in our ablation in Table \ref{tab:geod_ablation}. 

\begin{table}[htbp]
  \centering
  \caption{Geodesics Regression ablation experiments. Without using the straight line distance to help our network to predict the geodesic distances (Direct Prediction Method), we see 7\% poorer accuracy. }
    \begin{tabular}{|l|r|}
    \hline
    \textbf{Method} &
      \multicolumn{1}{l|}{\textbf{Error}}
      \bigstrut\\
    \hline
    LiteGE &
      2.3
      \bigstrut\\
    \hline
    Direct Prediction  &
      2.45
      \bigstrut\\
    \hline
    \end{tabular}%
  \label{tab:geod_ablation}%
\end{table}%
\subsection{UDF-PCA Representation Details}
\begin{figure}[htbp]
  \centering
  \includegraphics[width=0.8\columnwidth]{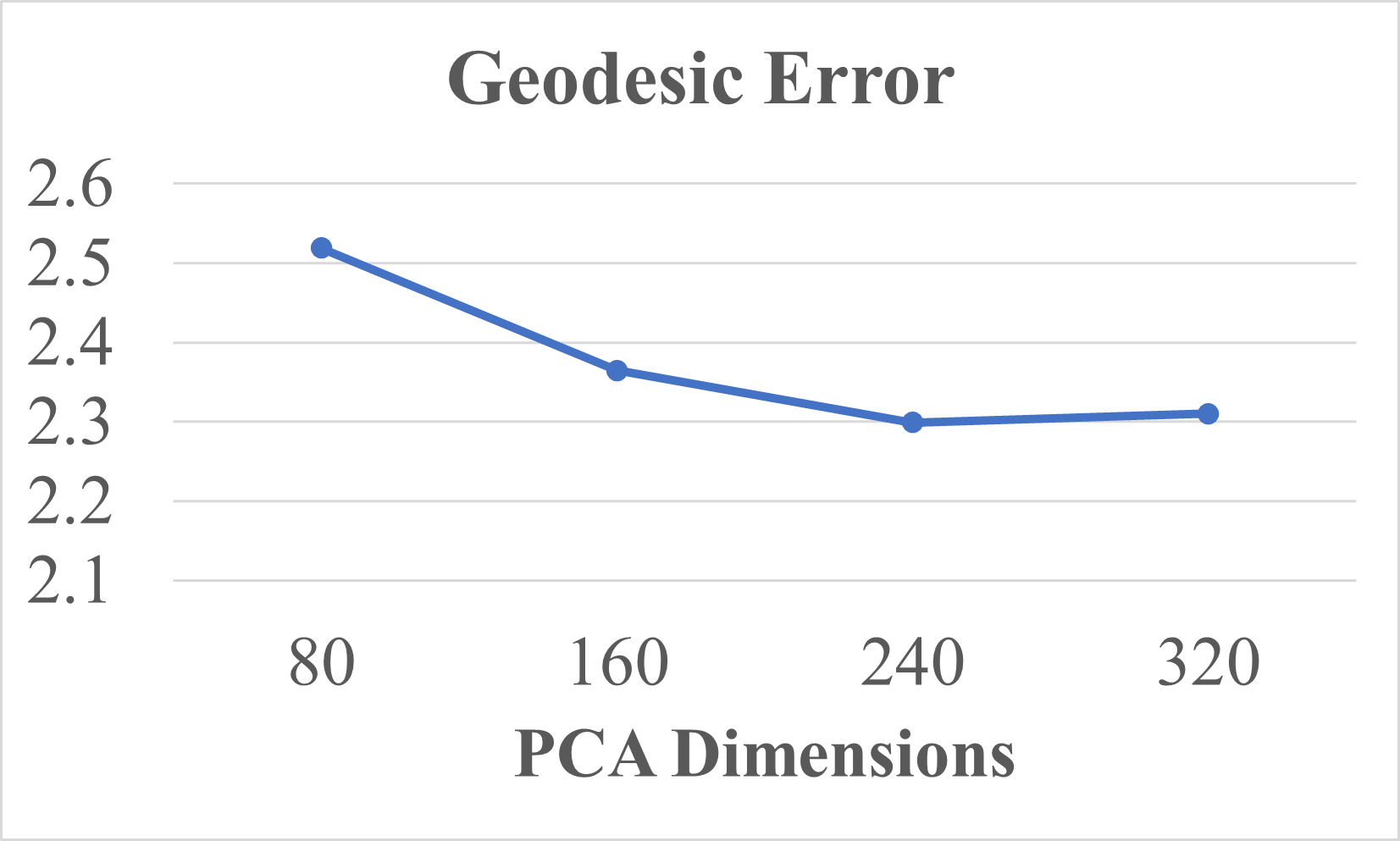} 
  \caption{This shows how the mean L1-error varies with varying number of PCA dimensions. As before, the geodesic ground truth are scaled uniformly so that its mean is 100. As we see, the error rate is pretty constant. Using lower PCA dimension can potentially lower our memory consumption.}
  \label{fig:geod_pcadim}
\end{figure}

Here, we chose approximately 20-25K important points around the shapes to represent each shape using unsigned distances vector. Using PCA, we reduced this representation to a 240-dimensional feature vector. Note that using different PCA dimension can still confer quite a good result as seen in Figure \ref{fig:geod_pcadim}. Using smaller PCA dimension does confer multiple-times memory reductions, but we have not explored this further yet. The epsilon threshold used for choosing the important points using the SMAL dataset is set to 0.0256, while for training using SMAL \cite{SMALCVPR:2017} and SURREAL dataset \cite{groueix20183d}, it is set to 0.0144. Then, for training using the Objaverse \cite{deitke2023objaverse} dataset, this threshold is set to 0.01. Also, note that during training, we normalize the 3D coordinates of query points using their mean and variance, but we keep the UDF-PCA features as it is. Our UDF-PCA representation is created from point clouds samples. We observed that normalizing the UDF-PCA features can significantly reduce the cosine similarity scores between 2 point clouds samples from the same shape, from around 0.9 - 1.0 to 0.0-0.4. This could potentially translate to poorer results. Nevertheless, we have not had time yet to do complete ablation on this.

\subsection{Geodesic Dataset}
\subsubsection{SMAL Geodesic Dataset.}
For our geodesic dataset on the SMAL models, we first separated the meshes there randomly to train, validation and test set. The geodesics distances ground truth training dataset is created by first sampling 80K mesh pairs randomly from the train set. Then, for each pairs, we computed geodesic distances between 10 random sources and 400 random destinations for each sources. This dataset is used to train both our shape matching module and our geodesic inference module on the SMAL dataset.
\subsubsection{SMAL and SURREAL Geodesic Dataset.}
For training using the SURREAL and SMAL dataset, after separating the meshes to train, validation and test set, we sample 40K mesh pairs from SURREAL train dataset, and 40K mesh pairs from SMAL train dataset. Then, for each pairs, we computed geodesic distances between 10 random sources and 400 random destinations for each sources that are used for training. 

\subsubsection{Objaverse Geodesic Dataset.}

The Objaverse geodesic dataset was created by initially sampling 4,000 meshes with fewer than 100,000 vertices. Larger meshes are not included to guarantee efficient geodesic computation. After splitting these meshes into training and validation sets, they were preprocessed using the Meshfix tool \cite{attene2010lightweight} to ensure successful geodesic computation using Fast-DGG \cite{fDGG}, discarding any corrupted files. For the remaining meshes in both the trainind and validation set, we computed all-pairs geodesic distances: specifically, between 300 randomly chosen vertices for meshes with $\geq300$ vertices, or between all vertices for smaller meshes. After we canonicalize those meshes using our canonicalization process explained in our paper, a critical filtering step was applied, retaining only meshes that were contained within a 3×3×3 bounding box centered at the origin. This rigorous procedure yielded approximately 3500 meshes for training and testing LiteGE, with around 100 million geodesic distances computed for them.

\subsubsection{Further Notes.}
During training, we scale the geodesic distance ground truth by 1.42. This is done to make the ground truth distances to have mean of around 0.86-0.9. Our training is inherited from a previous training setup, we want to keep the same ground truth data mean so that we won't have to change the learning rate again.

Here, we used the SMAL and SURREAL datasets as they contain popular animals and human models that are often used in many graphics and computer vision applications. The SMAL dataset is used particularly because it contains popular non-isometric animal models for testing our shape matching module. Meanwhile, the Objaverse dataset is used because it contains diverse 3D models that we can use to test LiteGE generalizability.

\section{Shape Matching Training and NN-Cache Details}
\subsection{Training Parameters}
Our training procedure for shape matching is similar with the training for geodesics. That is, we used L1 loss with Adams optimizer, 10 epochs and batch size 3072, with early stopping applied. Cosine annealing learning rate schedule that starts from $10^{-2}$ and ends with $10^{-4}$ is used. We trained the network to predict the average of geodesic distance when 2 points from shape X and shape Y are both lying on either shape X or shape Y. This strategy is based on our initial design. We didn't have time to try other designs yet.

For creating the UDF representation we used $\varepsilon$ threshold of 0.0196. Using PCA, we reduced this representation to a 400-dimensional feature vector. 

For our MLPs design, we used the same designs as our geodesic regression MLP's in the paper, but with 2x the number of neurons at each layer. All layers in the MLP used ReLU and batch normalization.

For our NN-cache, we used 3 level cache, with $N_1 = 60, N_2 = 650, M_1 = 65, M_2 = 60$ and $N_3$ being all vertices on the mesh when dealing with the 5K vertices remeshed and broken mesh models. When dealing with 10K anisotropic remeshed models, we also used the same configuration, but for simplicity $N_3$ is set to 5K only. The reason we used 5K samples in the final tier is for testing purposes. We only know the all pairs geodesics distances between 5K vertices samples, so we used as the anchors points. When testing point clouds models, we actually match the queries points from the point clouds of 1 shape to the 5K remeshed mesh model vertices in another shape. But, we used the 500-8K point cloud UDF-PCA representation to represent the mesh and 500-8K point cloud to showcase the strength of our representation. This is done for simplicity in our test. So, we also used this NN-cache configuration in our point cloud testing.

Then, for testing with template models with 4K vertices, we also used 3 level cache with $N_1 = 60, N_2 = 500, M_1 = 50, M_2 = 56$ and $N_3$ being all vertices on the mesh.

Finally, on poisson disk sampling for point cloud scaling \cite{corsini2012efficient}, when we deal with 8K point clouds, we used 500 poisson disk samples, while when we have 500 point clouds, we used 50 poisson disk samples. When we deal with broken meshes, we used 7K poisson disk samples. The target for average distance of the 1-nearest-neighbor in the poisson disk samples is set to 0.03835, when we deal with 8K point clouds, or 0.12121, for 500 point clouds, and 0.01181 when dealing with broken meshes.

Meanwhile, for the TNet network, we used 450 epochs for training the network on randomly rotated 2K point clouds from unit area meshes in the training dataset that has been centered to origin. Cosine annealing learning rate that starts from $10^{-3}$ and ends with $10^{-5}$ and batch size of 144 are used to train it. During TNet inference in the canonicalization process, we must center the point clouds and sample it from unit area meshes too. The rotation matrix computed using TNet is used to orient the meshes as well. When we deal with point clouds, as mentioned before, we scale them so that K poisson disk samples from the point cloud has the property that the average 1-nearest-neighbor distance is the same as the average 1-nearest-neighbor distance of K poisson disk samples from unit area mesh. This is done before using TNet for alignment.

Our TNet designs follows PointNet design that used a shared 3 layers MLP with 128 neurons at first layer, 512 neurons for the second one, 2048 neurons finally to predict features for each point and do global max pooling at the end to result in 2048 features representation of the point cloud. The 2048 features representation is then fed to MLP with 3 layers that has (2048, 512, 128) neurons hidden layers configuration to predict the 6D rotation vectors.
\subsection{Ablation Experiments}
\begin{table}[h]
  \centering
  \caption{Here, we presented our ablation experiment for shape matching. Due to time constraints, we evaluate only the first epoch accuracy. Here, "No Filter" means no rejection sampling is used to filter TNet output. Meanwhile, "Filtered (Best)" (resp. "Filtered (Worst)") means it is the best (resp. worst) network that we get after 8 training passes.  The error is measured in 100 x mean geodesic error on unit area mesh in the template meshes test data. The AUC metric is area under the curve for percentage of geodesic error below 0.0 - 0.1. }
    \begin{tabular}{|l|r|r|}
    \hline
    \multicolumn{1}{|c|}{\multirow{2}[4]{*}{\textbf{Method}}} &
      \multicolumn{2}{c|}{\textbf{First Epoch Accuracy}}
      \bigstrut\\
\cline{2-3}     &
      \multicolumn{1}{l|}{\textbf{AUC}} &
      \multicolumn{1}{l|}{\textbf{Error}}
      \bigstrut\\
    \hline
    Filtered (Best) &
      75.5 &
      3.05
      \bigstrut\\
    \hline
    Filtered (Worst) &
      70.6 &
      3.81
      \bigstrut\\
    \hline
    No Filter &
      54 &
      6.1
      \bigstrut\\
    \hline
    \end{tabular}%
  \label{tab:shapematch_ablation}%
\end{table}%

\begin{table}[htbp]
  \centering
  \caption{TNet Ablation results compared with using PCA axis alignment when training and testing using randomly rotated models. We presented 100 x mean geodesic error on unit area template dataset here.}
    \begin{tabular}{|l|r|}
    \hline
    \multicolumn{1}{|c|}{\multirow{2}[4]{*}{\textbf{Method}}} &
      \multicolumn{1}{c|}{\textbf{First Epoch Accuracy}}
      \bigstrut\\
\cline{2-2}   &
      \multicolumn{1}{l|}{\textbf{Error}}
      \bigstrut\\
    \hline
    LiteGE &
      3.05
      \bigstrut\\
    \hline
    PCA Axis Alignment &
      $\geq 15$
      \bigstrut\\
    \hline
    \end{tabular}%
  \label{tab:tnet_ablation}%
\end{table}%

When training the shape matching prediction network, we must do rejection sampling on the TNet output for training stability. As mentioned, we filtered the TNet aligned models that have errors larger than 0.66 radians. Without this, our accuracy drops significantly from around 70-75 AUC to 54 AUC in the template meshes test data, as seen in our ablation experiment in Table \ref{tab:shapematch_ablation}. We also see that it is important to do multiple training pass for 1 epoch and pick the best network as different training pass produce different results. This is seen in the first epoch accuracy of our method, that can differ between runs. In fact, the mean geodesic error of our method can differ by 25\% points between the best and worst training pass as seen in Table \ref{tab:shapematch_ablation}. 

TNet alignment is crucial in reducing data variability when we deal with a pair of shapes for shape matching. We presented our ablation on TNet network in Table \ref{tab:tnet_ablation}. Without TNet alignment, our first epoch error jumps from 3.0 to more than 15.0. This result underscores the importance of good normalization method for LiteGE.

\subsection{Spectral Meets Spatial Results}
To clear any doubts, we presented experiments on SMS here using different number of meshes and by using random remeshing strategy as data augmentation. It is seen in Table \ref{tab:sms_result_moremeshes_remeshed}, SMS results does not differ much when we used 6K meshes training data as compared when we used 1200 meshes as training data. In other words, SMS evaluation accuracy has been saturated after using 1200 meshes as training data. Thus, we can conclude that training with all of our training data that contains 12K meshes would likely result in only a negligible improvement in SMS accuracy.

Moreover, augmenting the training set with randomly remeshed models also fails to boost SMS accuracy on remeshed test data (Table \ref{tab:sms_result_moremeshes_remeshed}). The augmented training meshes (with 2K-10K vertices) were generated using isotropic explicit remeshing \cite{hoppe1993mesh} with random parameters.

\begin{table}[htbp]
  \centering
  \caption{SMS results on the 5K vertex isotropic-remeshed test models with different number of meshes used for training and data augmentation using randomly remeshed models. As before, 100 x mean geodesic error figure on unit area mesh is presented. AUC is calculated for geodesic errors with upper threshold of 0.2.}
    \begin{tabular}{|l|r|r|r|}
    \hline
    \textbf{Train Data Type} &
      \multicolumn{1}{l|}{\textbf{\#Meshes}} &
      \multicolumn{1}{l|}{\textbf{AUC}} &
      \multicolumn{1}{l|}{\textbf{Error}}
      \bigstrut\\
    \hline
    Randomly Remeshed &
      1200 &
      72.9 &
      7.51
      \bigstrut\\
    \hline
    Template &
      1200 &
      72.8 &
      7.55
      \bigstrut\\
    \hline
    Template &
      6000 &
      73.1 &
      7.44
      \bigstrut\\
    \hline
    \end{tabular}%
  \label{tab:sms_result_moremeshes_remeshed}%
\end{table}%


\bibliography{aaai2026}